\documentclass{llncs}
\usepackage[utf8]{inputenc}
\usepackage{amsmath}
\usepackage{textcomp}
\usepackage{graphicx}

\usepackage{hyperref}

\title{Synaptic Cleft Segmentation in Non-Isotropic Volume Electron Microscopy of the Complete \emph{Drosophila} Brain}
\titlerunning{Synaptic Cleft Segmentation in Non-Isotropic EM}
\author{%
Larissa Heinrich\inst{1}%
\and%
Jan Funke\inst{1,2}%
\and%
Constantin Pape\inst{1,3}%
\and%
Juan Nunez-Iglesias\inst{1}%
\and%
Stephan Saalfeld\inst{1}%
}
\institute{%
HHMI Janelia Research Campus%
\and%
Institut de Rob\`otica i Inform\`atica Industrial%
\and%
University of Heidelberg%
}

\begin{document}
\maketitle

\begin{abstract}
Neural circuit reconstruction at single synapse resolution is increasingly recognized as crucially important to decipher the function of biological nervous systems.  Volume electron microscopy in serial transmission or scanning mode has been demonstrated to provide the necessary resolution to segment or trace all neurites and to annotate all synaptic connections.

Automatic annotation of synaptic connections has been done successfully in near isotropic electron microscopy of vertebrate model organisms.  Results on non-isotropic data in insect models, however, are not yet on par with human annotation.

We designed a new 3D-U-Net architecture to optimally represent isotropic fields of view in non-isotropic data.  We used regression on a signed distance transform of manually annotated
synaptic clefts of the CREMI challenge dataset to train this model and observed significant improvement over the state of the art.

We developed open source software for optimized parallel prediction on very
large volumetric datasets and applied our model to predict synaptic clefts in a
50 tera-voxels dataset of the complete \emph{Drosophila} brain.  Our model
generalizes well to areas far away from where training data was available.
\end{abstract}

\section{Introduction}
Today, the neuroscience community widely agrees that the synaptic microcircuitry of biological nervous systems is important to understand what functions they implement.
The only currently available method to densely reconstruct all axons, dendrites, and synapses is volume electron microscopy (EM) as it provides a resolution sufficient to unambiguously separate them ($<15$\,nm per voxel, \cite{plazaal14}).  For EM \emph{connectomics}, several flavors of volume EM have been used \cite{briggmanb12}: Serial block face scanning EM (SBFSEM), in combination with focused ion beam milling (FIB-SEM), provides the highest isotropic resolution of $\sim5^3$\,nm per voxel and excellent signal to noise ratio but is relatively slow.  On the other end of the spectrum, serial section transmission EM (ssTEM) offers excellent lateral resolution, imaging speed, and signal to noise ratio but generates highly non-isotropic data with comparably poor axial resolution ($>35$\,nm per voxel).

A remarkable number of projects are currently under way to reconstruct the connectomes of various model organisms \cite{schlegelal17}, ranging from small invertebrate nervous systems like the larvae of \emph{Drosophila melanogaster} \cite{schneideral16,eichleral17} or \emph{Platynereis dumerilii} \cite{randelal15}, the adult \emph{Drosophila} \cite{takemuraal17,zhengal17}, to vertebrate models like the zebrafish larva \cite{hildebrandal16}, the retina of a mouse \cite{helmstaedteral13}, or the zebra finch HVC \cite{kornfeldal17}.

\section{Related work}
While many ongoing connectome reconstruction efforts still rely on manual annotation of synaptic contacts \cite{schneideral16,zhengal17}, automatic annotation of synaptic clefts from volume electron microscopy has been explored in recent years.  On vertebrate model systems, existing solutions perform comparably to trained human annotators on both isotropic \cite{kreshukal11,beckeral13,kreshukal15,staffleral17,dorkenwaldal17} and non-isotropic data \cite{kreshukal14,dorkenwaldal17}.
Synapses in the insect brain, however, are more complicated, and typically smaller than in vertebrates.  Accordingly, the performance on isotropic data is good \cite{kreshukal11,huangp14,plazaal14a}, but not yet satisfying on non-isotropic data (see CREMI leaderboard).\footnote{MICCAI Challenge on Circuit Reconstruction from Electron Microscopy Images (CREMI): \href{https://cremi.org}{https://cremi.org}}

The methods follow the general trend in computer vision.  Earlier approaches \cite{kreshukal11,kreshukal14,kreshukal15,beckeral13,plazaal14a} use carefully designed image features and train pixel classifiers using random forests or gradient boosting.  More recent approaches \cite{huangp14,staffleral17,dorkenwaldal17} train deep learning models to classify pixels or regions of interest as synapse candidates.  All approaches rely on sensible post-processing to filter false detections.

The CREMI challenge provides three volumes with ground truth for neuron
segmentation, synaptic clefts, and synaptic partner annotations in diverse
regions of ssTEM of the adult \emph{Drosophila} brain at 40\texttimes$4^2$\,nm
per voxel.  The challenge data includes typical artifacts for ssTEM preparations
such as missing sections, staining precipitate, or incorrect alignment.  To our
knowledge, it is the only existing challenge with secret test data that
enables unbiased comparison of synapse detection in non-isotropic EM of the
insect brain.  The evaluation metric for  synaptic cleft detection (CREMI score)
is the average of the average false positive distance (FPD) and the average
false negative distance (FND). The FPD is the distance of a predicted label to
the nearest true label, the FND is the distance of a true label to the nearest
predicted label.

\section{Methods}

\subsection{Training setup}

We corrected the serial section alignment errors present in the CREMI volumes
using elastic alignment with TrakEM2 \cite{SaalfeldAl12} and split each volume
into a training (75\%) and validation (25\%) subset, such that the statistics of
each subset are visually similar to the whole block.  We trained 3D-U-Nets
\cite{cicekal16} to predict a signed distance transform of binary synapse labels
using the TensorFlow library \cite{tensorflow2015}.  We used
Gunpowder\footnote{Gunpowder:
\href{https://github.com/funkey/gunpowder}{https://github.com/funkey/gunpowder}}
for batch loading, preprocessing, and training.  We made heavy use of
Gunpowder's support for data augmentations auch as transposing, intensity
variation, elastic deformations, and ssTEM-specific artifacts like missing or
noisy sections.  We believe that these augmentations are crucial for our network
to generalize well on large datasets without substantial engineering efforts.

As synaptic clefts are very sparse, we sample batches that contain synapses more frequently by rejecting batches without synapses with 95\% probability.  Additionally, we rebalance the loss with the frequency of positively annotated voxels to heavily penalize false negative predictions (unless otherwise stated).

We used Adam to minimize the L2 loss w.r.t. a signed Euclidean distance transform (SEDT) of the binary labels.  As the SEDT is not meaningful far away from synapses, we scaled it and applied a $\tanh$ nonlinearity that saturates between [-1,1]: $STDT = \tanh(SEDT/s)$.  Our experiments indicated that the scaling factor has little effect on performance (data not shown).  We chose $s=50$ as the default parameter.  Simple thresholding converts the predicted $STDT$ into binary labels.

\begin{figure}
 \includegraphics{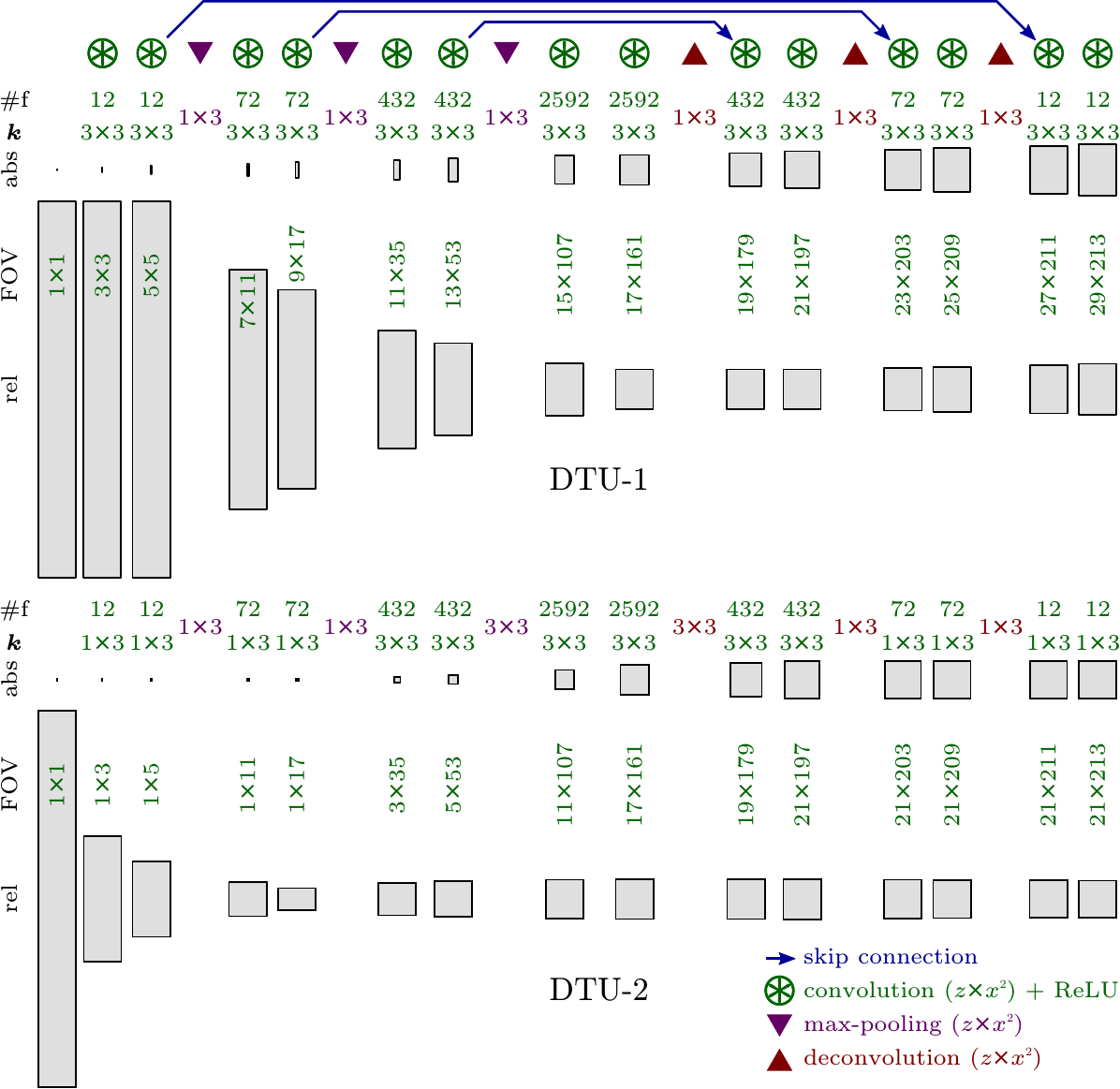}
 \caption{Comparison of the physical FOV in each layer of the 3D-U-Net
architectures DTU-1 and DTU-2.  The top row shows a graphical
representation of the general U-Net architecture.  The network consists of
modules of two convolutions and rectlinear units followed by max-pooling on the
encoding side, and deconvolution followed by two convolutions and rectlinear
units on the decoding side.  Kernel sizes ($\boldsymbol{k}$) are denoted as
$z$\texttimes$x$ as the $x$ and $y$ axes are isotropic.  The number of features
per convolutional layer (\#f) is increased by a factor of six after max-pooling
and decreased by a factor of six after deconvolution. In DTU-2, 3D-convolutions
are replaced by 2D-convolutions where the resolution is highly non-isotropic,
and 2D-max-pooling and deconvolution are replaced by 3D-max-pooling and
deconvolution where the resolution is near-isotropic.  The physical FOV in each layer, depicted as absolute (abs) and relative (rel) size
boxes, is therefore closer to isotropic than in DTU-1.}
 \label{fov}
\end{figure}

\subsection{Experiments}

\begin{figure}[t]
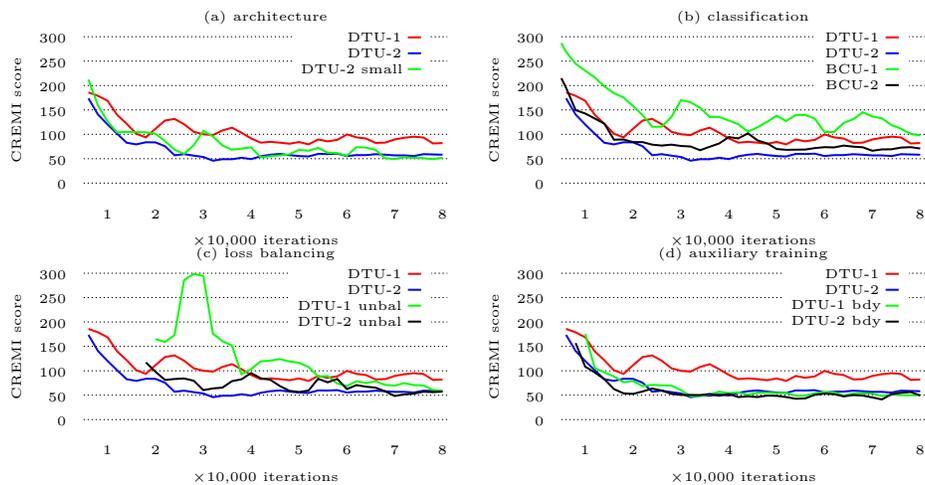

\vspace{-1.5cm}
\input{paper-dtu-gnuplottex-fig1.tex}

\input{paper-dtu-gnuplottex-fig2.tex}
\caption{Validation experiments.  All plots show mildly smoothed validation results sampled in intervals of 2,000 iterations using the CREMI score averaged over the validation set.  (a) shows that DTU-2 outperforms DTU-1, even if training blocks of the same size as for DTU-1 are used.  (b) shows that DTU-1 and DTU-2 trained for regression on the distance transform outperform the same architectures trained for binary classification. (c) shows that loss balancing makes training more robust.  (d) shows that auxiliary training for boundary distances improves performance on synaptic cleft detection.}
\label{fig:experiments}
\end{figure}

\subsubsection{3D-U-Nets benefit from isotropic fields of view (FOV).}The DTU-1 (distance transform U-Net) architecture is based on a design for neuron segmentation in non-isotropic ssTEM \cite{funkeal17} (see Fig.~\ref{fov}).  The physical FOV of this architecture is highly non-isotropic across a large number of layers.  Hypothesizing that an isotropic physical FOV would be beneficial to learn meaningful physical features, we tweaked the kernel sizes while retaining the overall design.  The `isotropic' network (DTU-2, see Fig.~\ref{fov}) is restricted to 2D convolutions in the first few levels and has isotropic kernels once the voxel size is nearly isotropic.  The encoding and decoding side are symmetric.  Fig.~\ref{fig:experiments} shows that DTU-2 consistently outperforms DTU-1.

DTU-2 has significantly fewer parameters than DTU-1.  This allows for a larger patch size (output size 23\texttimes218\texttimes218 as opposed to 56\texttimes56\texttimes56) which translates into a better estimate of the true gradient during stochastic gradient descent.  While this constitutes an additional advantage of DTU-2, we showed that it is not sufficient to explain its superior performance.  A smaller version of DTU-2, with output size 20\texttimes191\texttimes191, still outperforms DTU-1 (see Fig~\ref{fig:experiments}a).

At the time of writing, DTU-2 is first on the CREMI synaptic cleft detection challenge, followed by DTU-1 in second place.  Unlike the experiments shown in Fig.~\ref{fig:experiments}, those networks were trained for more iterations, on a curated version of the full CREMI ground truth.

\vspace{-1em}\subsubsection{Regression outperforms classification.}
Most deep-learning approaches for object detection in general, and synapse detection specifically \cite{dorkenwaldal17}, use a sigmoid nonlinearity and cross-entropy loss to predict a probability map for the object.  Inspired by the recent success of long-range affinities as an auxiliary loss for boundary detection \cite{leeal17}, we suspected that networks might generally benefit from being explicitly forced to gather information from a larger context.  With this assumption in mind, we trained the network to predict a distance rather than a probability map.  This approach turns the voxel-wise classification into a voxel-wise regression problem \cite{philippal15}.

In Fig.~\ref{fig:experiments}b, we compare the performance of probability map prediction using a sigmoid nonlinearity with binary cross entropy loss and the STDT prediction as shown before.  All other hyperparameters are the same and the maps are converted into binary labels with a non-tweaked threshold (i.e. 0.5 and 0, respectively).  For both network architectures, the CREMI score on the validation set improves when predicting the STDT.

\vspace{-1em}\subsubsection{Loss balancing is important.}
Rebalancing the loss is an important feature of the training pipeline (Fig.~\ref{fig:experiments}c).  In early iterations, the CREMI score cannot be properly evaluated as no voxel in the validation set is predicted to be above threshold, i.e. no synapses were detected.

\vspace{-1em}\subsubsection{Auxiliary training improves performance.}
As synaptic clefts are, by definition, located at cell boundaries, we conducted experiments to determine whether an auxiliary loss from predicting a distance map of cell boundaries boosts performance.  We added a second output channel to both DTU-1 and DTU-2 with an (unbalanced) L2 loss with respect to the STDT, now computed on the neuron labels. The two losses are weighed equally. Batch sampling is still done with respect to synaptic clefts.

Fig.~\ref{fig:experiments}d shows that both networks benefit from the auxiliary loss signal.  Interestingly, the effect is more significant for DTU-1. A careful evaluation of the boundary detection is beyond the scope of this work.

\subsection{Synaptic cleft prediction on the complete \textit{Drosophila} brain}

\begin{figure}[t]
 \includegraphics{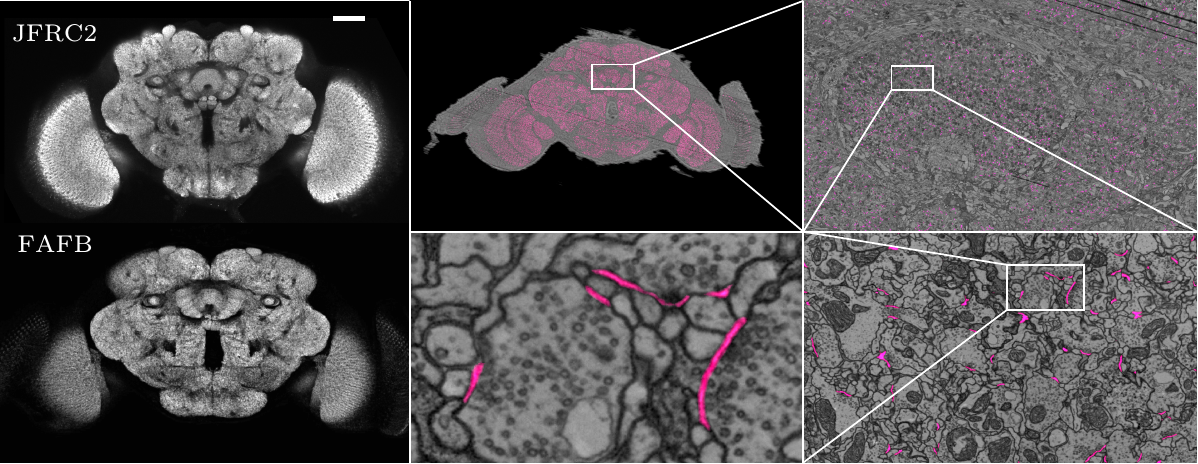}
 \caption{Synaptic cleft prediction on the complete adult \emph{Drosophila} brain. Left: Convolution of our predictions (FAFB) with a smooth PSF reproduces synaptic densities as visualized by fluorescent microscopy with the the nc82 antibody (JFRC2 template brain \cite{jenettal12}), scale bar 50\,\textmu{}m. Right: Examplary zoom-series into our DTU-2 synaptic cleft predictions overlayed over the FAFB volume.}
 \label{fig:examples}
\end{figure}

Prediction on large volumes can be performed in parallel on adjacent blocks.  Since our network was trained on valid input, the input block-size needs to be padded by the FOV of the network, i.e. while output blocks are adjacent and non-overlapping, input blocks overlap.

We converted the full adult fly brain (FAFB) volume \cite{zhengal17}\footnote{Available for download at \href{http://temca2data.org/}{http://temca2data.org/}} into a scale-pyramid using the the N5 data format\footnote{N5 specification: \href{https://github.com/saalfeldlab/n5}{https://github.com/saalfeldlab/n5}} on a shared filesystem.  We used N5 for both input and output because it enables parallel reading and writing of compressed volumetric blocks.  Prediction requires less memory than training because gradients do not need to be computed.  We found 71\texttimes650\texttimes650~voxels to be the maximum valid output block-size for the DTU-2 network that we could process on our NVIDIA Quadro M6000 GPUs with 12\,GB of RAM.  Using this increased block-size accelerated prediction by a factor of $\sim$2.5 compared to the block-size used for training.

The relevant biological sample covers only ~20\% of the FAFB volume.  We used ilastik \cite{sommeral11} to train a random forest classifier on scale-level 7 (downscaled by 13\texttimes128\texttimes128, i.e. $\sim$0.5$^3$\,\textmu{}m per voxel) that separates relevant biological sample from background.  Only output blocks that intersect with this mask were considered for prediction.  This valid set of blocks has a volume of $\sim$50~tera-voxels, the entire FAFB volume including background contains $\sim$213~tera-voxels.

We distributed the list of output blocks over 48 GPUs.  For each GPU, we used
Dask \cite{rocklin15} to load, preprocess, and store image blocks while the GPU
performed prediction, achieving greater than 90\% GPU utilization\footnote{Parallel prediction framework: \href{https://github.com/saalfeldlab/simpleference}{https://github.com/saalfeldlab/simpleference}}.  Our average
prediction speed was $\sim$3~mega-voxels per second and GPU, i.e. prediction
of the complete volume was finished in less than five days.

The quality of predictions across the entire volume was consistent with our results on CREMI (see Fig.~\ref{fig:examples}).  Even in areas with different characteristics than the CREMI training volumes (such as the lamina), synaptic cleft predictions are mostly correct and consistent with our expectations.  Predictions are correctly missing in axonal tracts and in the cortex.  We produced a simulation of an nc82 labeled confocal image by applying a large non-isotropic Gaussian PSF to our predictions and visually compared the result with the JFRC2 template brain \cite{jenettal12} (see Fig.~\ref{fig:examples}).  Accounting for that the two volumes stem from different individuals and have not been registered, our predictions convincingly reproduce the synaptic density distribution as visualized with the  nc82 antibody.

\section{Conclusion}
In this paper, we described a significant improvement over the state of the art in detection and segmentation of synaptic clefts in non-isotropic ssTEM of the insect nervous system.  We designed a 3D-U-Net architecture and training scheme that is particularly well suited to account for the non-isotropy in ssTEM data and the sparsity of synapses.  We trained this architecture by regression on a signed distance transform of manually annotated synaptic clefts of the publicly available CREMI challenge.  We showed that our new architecture compares favorably to a previously described architecture for the same data despite exposing fewer training parameters.  We developed an optimized framework for parallel prediction on very large volumetric data and achieved a prediction throughput of $\sim$3~mega-voxels per second and GPU.  This efficiency enabled us to predict all synaptic clefts in the 50~tera-voxels full adult \emph{Drosophila} brain \cite{zhengal17} in less than five days.  We made our code publicly available as open source under a permissive license\footnote{CNNectome: \href{https://github.com/saalfeldlab/cnnectome}{https://github.com/saalfeldlab/cnnectome}}.

\bibliographystyle{splncs03}
\bibliography{references.bib}

\end{document}